\documentclass[12pt]{article}
\usepackage{sbc-template}
\usepackage{graphicx,url}
\usepackage{xcolor,float}
\usepackage{float}   
\usepackage{hyperref}
\usepackage{amssymb}
\usepackage{scalefnt}
\usepackage{acronym}

\usepackage{scalerel}

\usepackage{algorithm}
\usepackage{algpseudocode}
\usepackage{amssymb}

\hypersetup{
  colorlinks   = true, %Colours links instead of ugly boxes
  urlcolor     = black, %Colour for external hyperlinks
  linkcolor    = black, %Colour of internal links
  citecolor   = black %Colour of citations
}

\usepackage{mathtools}

\title{Evaluating the Potential of Federated Learning \\for Maize Leaf Disease Prediction}

\author{Thalita Mendonça {Antico}\inst{1}, Larissa F. {Rodrigues Moreira}\inst{2}, Rodrigo {Moreira}\inst{1}}
\address{Institute of Exact and Technological Sciences (IEP) \\Federal University of Viçosa (UFV), Rio Paranaíba, MG, Brazil
\nextinstitute
Faculty of Computing (FACOM) \\ Federal University of Uberlândia (UFU), Uberlândia, MG, Brazil
  \email{\{thalita.antico,rodrigo\}@ufv.br,\{larissarodrigues\}@ufu.br}}

\begin{document} 

\maketitle

\begin{abstract}
  The diagnosis of diseases in food crops based on machine learning seemed satisfactory and suitable for use on a large scale. The Convolutional Neural Networks (CNNs) perform accurately in the disease prediction considering the image capture of the crop leaf, being extensively enhanced in the literature. These machine learning techniques fall short in data privacy, as they require sharing the data in the training process with a central server, disregarding competitive or regulatory concerns. Thus, Federated Learning (FL) aims to support distributed training to address recognized gaps in centralized training. As far as we know, this paper inaugurates the use and evaluation of FL applied in maize leaf diseases. We evaluated the performance of five CNNs trained under the distributed paradigm and measured their training time compared to the classification performance. In addition, we consider the suitability of distributed training considering the volume of network traffic and the number of parameters of each CNN. Our results indicate that FL potentially enhances data privacy in heterogeneous domains.
\end{abstract}

%--A--%
\acrodef{AI}{Artificial Intelligence}
%--B--%

%--C--%
\acrodef{CNN}{Convolutional Neural Network}
\acrodef{CNNs}{Convolutional Neural Networks}
%--D--%

%--E--%

%--F--%
\acrodef{FL}{Federated Learning}
%--G--%

%--H--%

%--I--%
\acrodef{IEEE}{Institute of Electrical and Electronics Engineers}

%--P--%
\acrodef{PCA}{Principal Components Analysis}

\section{Introduction}\label{sec:introduction}

The rapid evolution of technologies in parallel with the increasingly sophisticated demands of users has generated unprecedented amounts of data. This large amount of data contains strategic information and knowledge whose extraction requires sophisticated computational methods. The extraction of dataset features can enable the construction of machine learning models to work in domains such as health, economics, and agriculture~\cite{Naeem2022}. 

Primarily, agriculture represents one of the main economic activities in the world, maize being the crop most produced in the world, leading Brazil to be the second largest exporter~\cite{Aragao2021}. However, the maize crops are susceptible to certain leave diseases, affecting production and reducing quality and productivity~\cite{Roese2020}. To overcome this limitation, \ac{AI} techniques allow building intelligent applications to support crop disease diagnosis~\cite{Shaikh2022}.

Machine learning paradigms, mainly supervised learning, require data to be centrally and locally available in the training phase. However, computing and storage resources are widely spread over multiple regions, and organizations require distributed training strategies to employ these resources~\cite{Li2020_FL}.

However, heterogeneously distributed computing resources impose privacy and security challenges. Therefore, the \ac{FL} paradigm emerges as a technique that exploits the potential of distributed resources by collaboratively training a machine pairing model~\cite{Liu2022}. The \ac{FL} has already shown promise in recent applications in different fields such as medicine~\cite{Dayan2021, Adnan2022}, industrial engineering~\cite{Hu2018, Mowla2020}, and mobile devices~\cite{Bochie2021}.

This paper sheds light on the potential of \ac{FL} as a suitable distributed machine learning approach for maize leaf diseases. As far as we know, this paper inaugurates the use and evaluation of \ac{FL} applied in diagnosing diseases in maize leaves based on \ac{CNNs} and employing a centralized learning paradigm, where the data must be present in a central server to support the training phase. In the \ac{FL} approach, the data could be spread over the training clients, which works distributed and asynchronously targeting to build a central model.

In a nutshell, our main contributions are: (1) a short state-of-the-art survey towards \ac{FL} applied to agriculture challenge; (2) a numerical and practical evaluation of \ac{FL} method for maize leaves disease; (3) a traffic volume evaluation of \ac{CNN} distributed trained; (4) a performance and behavior evaluation \ac{CNNs} distributed trained. 

The remaining of the paper is organized as follows: Section~\ref{sec:related_works} presents related work and highlights the contribution of our approach. Section~\ref{sec:proposed_method} describes our proposed method. The performance evaluation is presented in the Section~\ref{sec:performance_evaluation}. In Section~\ref{sec:results_and_discussion}, we present and discuss the obtained results. Finally, Section~\ref{sec:conclusion} concludes the paper.

\section{Related Work}\label{sec:related_works}

Maize leaf disease identification based on the image is a research field with many approaches developed over the years~\cite{Sharma2021}. \cite{DeChant2017}~proposed the automatic identification of northern maize leaf blight using a pipeline of \ac{CNNs}. \cite{Lin2018}~designed a multichannel \ac{CNN} with cross-connection, pooling, and normalization schemes. They evaluated five types of images collected from maize planting in China. 

\cite{Zhang2018}~enhanced the Cifar10 and GoogLeNet \ac{CNNs} tuning parameters, modifying the pooling combinations, adding dropout operations, and rectified linear unit functions to speed the training step. \cite{Alehegn2019}~proposed an approach based on color, texture, and morphological features to classify the Ethiopian maize diseases dataset using Support Vector Machine (SVM).

\cite{Priyadharshini2019}~proposed modifications in the LeNet architecture and applied a preprocessing step in the images using \ac{PCA} whitening. \cite{Panigrahi2020}~extracted color, shape, and textures features and evaluated several supervised classifiers to identify maize leaf disease, including Naïve Bayes, Decisoin Tree, K-Nearest Neighbors (KNN), SVM, and Random Forest.  

\cite{Waheed2020}~applied a hyperparameter tuning based on grid search to improve the classification performance of DenseNet. \cite{Rocha2020} proposed the hyperparameter optimization based on the Bayesian method and evaluated AlexNet, ResNet50, and SqueezeNet \ac{CNNs} trained with data augmentation and transfer learning. Similarly, \cite{Subramanian2022_2}~and \cite{Subramanian2022}~also applied Bayesian optimization and tested a dataset of images obtained from different sources.

\cite{Haque2022}~evaluated three CNN architecture inspired by InceptionV3 to identify maize leaf diseases in images collected in India. They used pre-trained models and applied data augmentation using rotation and brightness enhancement methods. \cite{Amin2022}~proposed a fusion of \ac{CNNs}, considering EfficientB0 and DenseNet121 models to discover the complex feature of the infected maize leaf images.

\cite{Moreira2022}~proposed a low-cost and green-friendly architecture called AgroLens to support Smart Farm applications in environments with poor connectivity. They compared the classification performance of \ac{CNNs} to handle maize leaf disease diagnosing supported by edge-compute devices.

In Table~\ref{tab:survey} we compared the recent contributions available in the literature in the scope of solutions to support maize leaf diagnosis based on images. The ``Approach'' column refers to the state-of-the-art proposals. The ``Dataset'' column indicates the dataset image considered. The ``ML Algorithms'' column indicates the algorithm utilized. The ``Training Place'' column characterizes if the solution was trained centralized or distributed. The ``Evaluation Metric'' indicates the classification performance considered. We observed in our short survey that all approaches considered accuracy evaluation metrics, and the accuracy obtained for each approach is indicated in the ``Accuracy'' column. Finally, the ``Privacy Support'' column characterizes the solution's ability to support data privacy. Classification performance is the main strength of previous studies based on machine learning algorithms. However, these studies do not investigate distributed training. Thus, to overcome this gap, we propose an approach based on \ac{FL}. Also, our method is more robust in addressing critical issues such as data privacy and security.

\begin{table}[!htbp]
\renewcommand{\arraystretch}{1.3}\scalefont{2}
\centering
\caption{Short State-of-the-Art Survey.}
\label{tab:survey}
\resizebox{\textwidth}{!}{%
\begin{tabular}{lllclcc}
\hline \hline
\multicolumn{1}{c}{\textbf{Approach}} & \multicolumn{1}{c}{\textbf{Dataset}} & \multicolumn{1}{c}{\textbf{ML Algorithm}} & \textbf{Traning Place} & \multicolumn{1}{c}{\textbf{\begin{tabular}[c]{@{}c@{}}Evaluation \\ Metric\end{tabular}}} & \textbf{Accuracy (\%)} & \textbf{\begin{tabular}[c]{@{}c@{}}Privacy  \\ Support\end{tabular}} \\ \hline
\cite{DeChant2017} & Northern Leaf Blight 2016 & CNN & Centralized & \begin{tabular}[c]{@{}l@{}}Accuracy Precision, \\ Recall, F1-Score\end{tabular} & 96.7 & $\times$ \\ \hline
\cite{Lin2018} & Own collected dataset & CNN Multichannel & Centralized & Accuracy & 92.31 & $\times$ \\ \hline
\cite{Zhang2018} & PlantVillage and Google websites & CNN Cifar10 and GoogLeNet & Centralized & Accuracy & 98.4 & $\times$ \\ \hline
\cite{Alehegn2019} & Own collected dataset & SVM & Centralized & Accuracy & 95.63 & $\times$ \\ \hline
\cite{Priyadharshini2019} & PlantVillage & CNN LeNet & Centralized & Accuracy & 97.89 & $\times$ \\ \hline
\cite{Panigrahi2020} & PlantVillage & \begin{tabular}[c]{@{}l@{}}Naive Bayes, Decision Tree\\ KNN, SVM, Random Forest\end{tabular} & Centralized & Accuracy & 79.23 & $\times$ \\ \hline
\cite{Waheed2020} & Different sources & Dense CNN (DenseNet) & Centralized & Accuracy & 98.06 & $\times$ \\ \hline
\cite{Rocha2020} & PlantVillage & \begin{tabular}[c]{@{}l@{}}\ac{CNNs} with Bayesian Optimization \\ AlexNet, ResNet-50, SqueezeNet\end{tabular} & Centralized & \begin{tabular}[c]{@{}l@{}}Accuracy Precision, \\ Recall, F1-Score\end{tabular} & 97 & $\times$ \\ \hline
\cite{Subramanian2022_2} & \begin{tabular}[c]{@{}l@{}}PlantVillage, Kaggle, \\ and  Google websites\end{tabular} & CNN VGG16 & Centralized & \begin{tabular}[c]{@{}l@{}}Accuracy Precision, \\ Recall, F1-Score\end{tabular} & 99.21 & $\times$ \\ \hline
\cite{Subramanian2022} & \begin{tabular}[c]{@{}l@{}}PlantVillage, Kaggle, \\ and  Google websites\end{tabular} & \begin{tabular}[c]{@{}l@{}}\ac{CNNs} \\ VGG16, ResNet50,\\ InceptionV3, Xception\end{tabular} & Centralized & \begin{tabular}[c]{@{}l@{}}Accuracy Precision, \\ Recall, F1-Score\end{tabular} & 99 & $\times$ \\ \hline
\cite{Haque2022} & Own collected dataset & CNN InceptionV3 & Centralized & \begin{tabular}[c]{@{}l@{}} Accuracy Precision, \\ Recall, F1-Score, Loss\end{tabular} & 95.99 & $\times$ \\ \hline
\cite{Amin2022} & PlantVillage & \begin{tabular}[c]{@{}l@{}}Fusion of \ac{CNNs} \\ EfficientNetB0, DenseNet\end{tabular} & Centralized & \begin{tabular}[c]{@{}l@{}}Accuracy Precision, \\ Recall, F1-Score\end{tabular} & 98.56 & $\times$ \\ \hline
\cite{Moreira2022} & PlantVillage & \begin{tabular}[c]{@{}l@{}}Embedded \ac{CNNs} \\ AlexNet, DenseNet, SqueezeNet\end{tabular} & Centralized & \begin{tabular}[c]{@{}l@{}}Accuracy Precision, \\ Recall, F1-Score\end{tabular} & 98.45 & $\times$ \\ \hline
\textbf{Our Approach (2022)} & \textbf{PlantVillage} & \textbf{\ac{CNNs} with \ac{FL}} & \textbf{Distributed} & \textbf{\begin{tabular}[c]{@{}l@{}}Accuracy Precision, \\ Recall, F1-Score\end{tabular}} & \textbf{97.29} & \textbf{\checkmark} \\ \hline \hline
\end{tabular}%
}
\end{table}
\section{Proposed Method}\label{sec:proposed_method}

Traditional machine learning requires the dataset to be locally available to feed the model in the training process. Some datasets require access control and privacy, such as sensitive data from hospitals, governments, businesses, banks, and others~\cite{Mugunthan2020}. \ac{FL} is a machine learning approach that does not require the data used in the supervised training process to be on the central server, leveraging data privacy as it can be processed locally~\cite{Truex2019, Zhang2021}.

\begin{figure}[!htbp]
  \centering
  \includegraphics[width=0.55\textwidth]{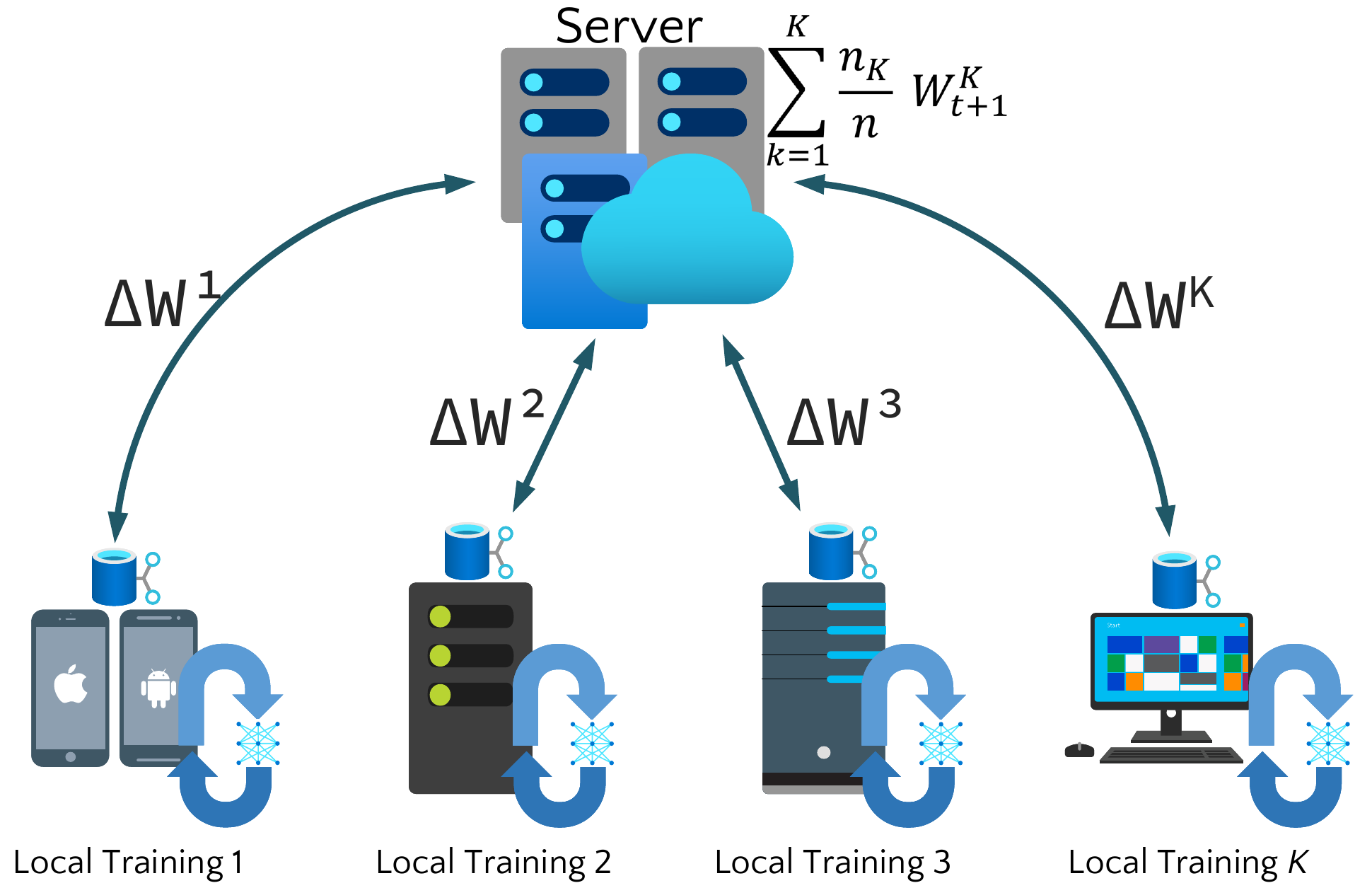}
  \caption{Federated Learning Rationale.}
  \label{fig:federated_learning_rationale}
\end{figure}

Figure~\ref{fig:federated_learning_rationale} depicts the distribution of the training workload among clients that could join the hierarchy asynchronously. In this machine learning paradigm, the client locally trains the model with its data, protecting it in terms of privacy, and submits only the weights of the parameters of the used machine learning model to the central server. This learning paradigm supports heterogeneous client's hardware capacity making \ac{FL} a dynamic approach.

When receiving the weights from the trained machine learning model, the Server performs a mathematical merge operation on the weights of each model reported by the clients who trained locally on the dataset. Thus, the Server maintains a single machine learning model containing weights referring to local training, not requiring the Server to access specific data for each location, protecting the principles of privacy on sensitive data. There are several mathematical operations in the community to merge the weights that the central server can adopt, some known approaches are: \textit{FedAvg}~\cite{Singh2017}, \textit{P-FedAvg}~\cite{Zhong2021}, \textit {FedProx}~\cite{Li2020}, \textit{EdgeFed}~\cite{Ye2020}, \textit{SimFL}~\cite{Li_Wen_He_2020} and others~\cite{Li2021}.

This paper presents a performance evaluation of \ac{CNNs} trained under the distributed paradigm. Thus, using \ac{FL} we consider one (1) central server and $K$ clients that connect to the server via the \ac{IEEE} 802.3 protocol~\cite{brooks2001ethernet}. Each customer receives a percentage proportional to the number of customers in the PlantVillage\footnote{Available in: \url{https://github.com/spMohanty/PlantVillage-Dataset}} dataset, specifically from the maize culture, categorized into four classes included common rust (1192 images), gray leaf (513 images), northern leaf blight (985 images), and healthy leaves (1162 images)~\cite{Mohanty2016}. Thus, these clients using \ac{CNNs} start the local training process, where each client trains \ac{CNN} according to their learning rate, epochs, and others.

\begin{figure}[!htbp]
  \centering
  \includegraphics[width=0.79\textwidth]{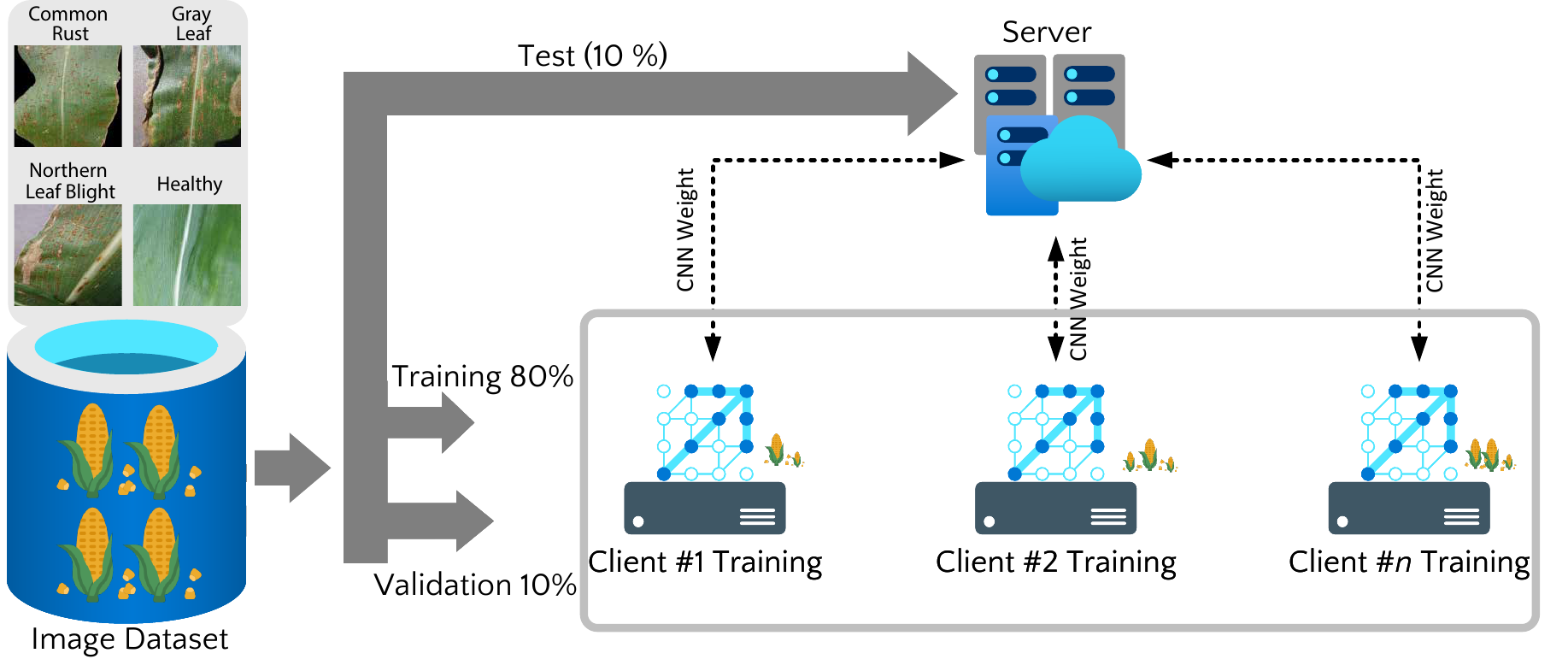}
  \caption{Proposed Method.}
  \label{fig:paper_method}
\end{figure}

Figure~\ref{fig:paper_method} illustrates the method used in this paper, which consists of using \ac{FL} as a distributed training paradigm. The maize leaf dataset feeds \ac{CNN}, which runs locally on each client. When the client ends its training cycle, it uploads the weights to the central \ac{CNN} model hosted by the central server.

We onboarded the federated clients with five \ac{CNNs}: AlexNet, SqueezeNet, ResNet-18, VGG-11, and ShuffleNet. These \ac{CNNs} were chosen based on their success in previous classification tasks. 

AlexNet~\cite{Krizhevsky2012} was the champion of the ImageNet Large Scale Visual Recognition Competition (ILSVRC) in 2012. It comprises five convolutional layers, three pooling layers, two fully connected layers, and a single Softmax output layer. Also, it uses dropout connections to reduce the overfitting and REctified Linear Unit (ReLU) activation function. 

SqueezeNet~\cite{Iandola2016} is a lightweight CNN architecture with approximately 50 times fewer parameters than AlexNet. It comprises convolutional layers, pooling layers, and fire layers. The fire layers perform the functions of fully connected layers and dense layers, which are not present in SqueezeNet. In addition, SqueezeNet applies compression techniques to reduce the number of parameters.

ResNet-18~\cite{He2016} is a residual network designed to deal with vanishing gradient problems using identity connection. The identity connection provides a direct pathway for the gradient without a weight layer and does not change the value of the gradient. In addition, ResNet is composed of two dominant blocks: identity and convolutional. The identity block is the standard block used in ResNet and performs identity mapping. The convolutional block reduces the dimension of the activation. In our study, we considered a ResNet with 18 layers.

VGG-11~\cite{Simonyan2014} is one of the \ac{CNNs} belonging to the Visual Geometry Group (VGG) models designed to replace the larger filters with large sequences of convolutional filters of size 3 $\times$ 3. We considered VGG with 11 weight layers: eight convolutional layers and three fully connected layers.

Finally, ShuffleNet~\cite{Ma2018} is a lightweight CNN that applies grouped depthwise separable convolutions to generate feature maps. Also, this CNN included a shuffle operation that shuffles the channels within feature maps is introduced as the second stage of convolution.

We validated the performance of each \ac{CNN} considering the set of tests divided according to the percentage of $80\%$ images for training, $10\%$ for validation, and $10\%$ for testing. The process of merging the weights of the \ac{CNNs} in this paper considered the \textit{FedAvg}~\cite{Singh2017}. The Algorithm~\ref{alg:fed_avg} represents the steps of \textit{FedAvg} divided into two different executions, the one executed by the Server and the one executed by the Client. In the Server execution, the model is initialized (line 2), and the highest value between the $C$ product is taken for each execution round. $K$, where $C$ is the participation ratio of the clients at each round and $K$ refers to the number of all clients (line 4).

\begin{algorithm}[!htbp]
\renewcommand{\arraystretch}{0.8}\scalefont{0.8}
\caption{\textit{Federated Averaging}. \textit{K} clients are indexed by \textit{k}; B is the size of the local \textit{minibach}, \textit{E} is the local number of epochs, and \textit{n} is the learning rate}\label{alg:fed_avg}
\begin{algorithmic}[1]
\State \textbf{Server Runs}\;
\State Initializes the Model $W_{0}$\;
\For{each round $t$ = 1, 2, ..., }
    \State m $\gets max(C.K, 1)$\;
    \State $S_{t}$ $\gets$ (random set of \textit{m}\; clients)\; 
        \For{for each \textit{K} $\in S_{t}$ in parallel do}
            \State $w_{t+1}^{k}$ $\gets$ UpdateClient(k, $w_{t}$)\;
            \State $w_{t + 1} \gets \sum_{k=1}^{k} \frac{n_{k}}{n}w_{t+1}^{k}$\;
        \EndFor
\EndFor
\State \textbf{UpdateClient}($K, w$)\;
\State $\beta \gets$ (divide $P_{k}$ in \textit{batches} of size B)\;
\For{each local epoch $i$ from 1 to E}
\State $w \gets w - n\triangledown l(w;b)$\;
\EndFor
\State return $w$ to server\;
\end{algorithmic}
\end{algorithm}

Thus, it chooses a random set of customers $S_{t}$ for training (line 5). The innermost loop (line 6) refers to the parallel execution of local, asynchronous, and distributed training that Clients trigger. As Clients train their models locally, they send their $w$ weights to the Server via \textit {UpdateClient} (line 7), making the Server update the weights of the general model (line 8) by the arithmetic mean mathematical operation. From the Client's perspective, the execution of \textit{ClientUpdate} consists of dividing the dataset into mini-batches of size B (line 12) and interactively, the minibatch $\beta$ feeds the local model in the training process of each Client (line 13). When the training finishes, its weight $w$ will be sent to Server to improve the global model (line 16).

We used the dataset division in the Independent and Identically Distributed (IID) model in our federated learning approach to classifying diseases in maize leaves. In this division, each client runs the training on a dataset part distributed evenly across all of them.

\section{Performance Evaluation}\label{sec:performance_evaluation}

We built an experimental scenario to validate the suitability of \ac{FL} training for different \ac{CNNs}. The experimental scenario consists of a machine with a 3.00 GHz Intel Processor i5 processor, 32 GB of RAM, an NVIDIA GeForce GTX 1080 Ti GPU, and Ubuntu 16.04.7 LTS operating system. We used the PyTorch framework (version 1.6) for the deep learning modules and CUDA version 8.0 and cuDNN 6.0. The \ac{FL} framework used in the performance evaluation was \textit{FedLab}~\cite{Zeng2021}.

On this machine, we run $k$ instances of Clients and 1 Server connected by changing the weights of \ac{CNN} at each training stage via socket. The parameters of the experimental scenario are according to Table~\ref{tab:experimental_parameters}.

\begin{table}[!htbp]
\centering
\caption{Experimental Setup Parameters.}
\label{tab:experimental_parameters}
\resizebox{\textwidth}{!}{%
\begin{tabular}{ccccccccccc}
\cline{2-11}
\multicolumn{1}{l}{} & \textbf{World Size} & \textbf{Rank} & \textbf{Epoch} & \textbf{Learning Rate} & \textbf{Batch Size} & \textbf{Momentum} & \textbf{Classes} & \textbf{Workers} & \textbf{IP} & \textbf{Port} \\ \hline \hline
Client \textit{k}             & [3, 5]              & \textit{k}    & 50             & 0.001                  & 32                  & 0.9               & 4                & 4                & 127.0.0.1   & Any           \\ \hline
Server               & [3, 5]              & 1             & 50             & -                      & 32                  & -                 & 4                & 4                & 127.0.0.1   & 3002          \\ \hline \hline
\end{tabular}%
}
\end{table}

At the end of each run, according to the parameters in Table~\ref{tab:experimental_parameters}, \textit{FedLab} generates performance metrics for each \ac{CNN}. Accuracy, Precision, Recall, and F1-Score are used to assess the classification performance. These metrics considered the indices based on the number of true positives (TP), true negatives (TN), false positives (FP), and false-negative (FN)~\cite{Tan2018}.

\begin{itemize}

    \item Accuracy: is the hits of the classifier as whole (Eq.~\ref{eq:acc}).
    {\small{
    \begin{equation}
    \label{eq:acc}
    \centering 
    Accuracy = \frac{T_{P} + T_{N}}{T_{P} + T_{N} + F_{P} + F_{N}}
    \end{equation}}}
    
    \item Precision: is the hit rate per class for positive cases (Eq.~\ref{eq:precision}).
    
    {\small{
    \begin{equation}
        \label{eq:precision}
        \centering 
        Precision = \frac{T_{P}}{T_{P} + F_{P}}
    \end{equation}}}

    \item Recall: is the rate of a relevant sample being classified correctly (Eq.~\ref{eq:recall}).
    
    {\small{
    \begin{equation}
        \label{eq:recall}
        \centering 
        Recall = \frac{T_{P}}{T_{P} + F_{N}}
    \end{equation}}}

    \item F1-Score: is the harmonic mean of Recall and Precision (Eq.~\ref{eq:f1}).
    
    {\small{
    \begin{equation}
        \label{eq:f1}
        \centering 
        F1\!\!-\!Score  = 2 \times \frac{Recall \times Precision}{Recall + Precision}
    \end{equation}}}    
\end{itemize}

\section{Results and Discussion}\label{sec:results_and_discussion}

We carried out experiments validating the performance of each \ac{CNN} trained with the \ac{FL} paradigm. For each \ac{CNN}, we run ten ($10$) executions with a random seed to assess the maximum performance that \ac{CNN} can achieve. Thus, according to Table~\ref{tab:results}, we found that the \ac{CNNs} with the highest accuracy were VGG-11 following AlexNet, with $97.29\%$ and $96.87\%$, respectively. In addition, our results demonstrate that AlexNet was \ac{CNN} that required the least time for training. Although the VGG-11 reached a high accuracy, this model demanded the most significant time in the training process, being unsuitable in \ac{FL} scenarios where the training time is essential.

We measured the standard deviation of the ten (10) experiments for each \ac{CNN}, recording it in Table~\ref{tab:results} after the $\pm$ symbol. These experiments were run without seed locking to measure the variability of the results. We consider in our experiments the training time of \ac{CNNs} considering the parameters of Table~\ref{tab:experimental_parameters}.

\begin{table}[!htbp]
\centering
\caption{\ac{CNN} Performance Comparison.}
\label{tab:results}
\resizebox{\textwidth}{!}{%
\begin{tabular}{lcccccc}
\hline \hline
\multicolumn{1}{c}{\textbf{CNN}} & \textbf{Accuracy (\%)} & \textbf{Precision (\%)} & \textbf{Recall (\%)} & \textbf{F1-Score (\%)} & \textbf{Loss} & \textbf{Training Time (min)} \\ \hline \hline
\textbf{AlexNet} & \textbf{96.87} $\pm$ \textbf{0.34} & \textbf{94.75} $\pm$ \textbf{0.06} & \textbf{95.25} $\pm$ \textbf{0.05} & \textbf{94.75} $\pm$ \textbf{0.15} & \textbf{0.96} & \textbf{16.23} \\
SqueezeNet & 96.54 $\pm$ 0.83 & 92.5 $\pm$ 0.67 & 94.25 $\pm$ 0.72 & 95.25 $\pm$ 1.01 & 1.22 & 25.55 \\
ResNet-18 & 94.86 $\pm$ 2.17 & 94.63 $\pm$ 0.18 & 94.80 $\pm$ 0.42 & 94.71 $\pm$ 0.29 & 2.84 & 27.95 \\
\textbf{VGG-11} & \textbf{97.29 $\pm$ 0.18} & \textbf{95.94 $\pm$ 0.21} & \textbf{96.59 $\pm$ 0.34} & \textbf{96.24 $\pm$ 0.27} & 6.45 & 78.78 \\
ShuffleNet & 75.65 $\pm$ 2.00 & 57.84 $\pm$ 0.61 & 64.27 $\pm$ 1.97 & 59.59 $\pm$ 1.96 & 16.25 & 19.99 \\ \hline \hline
\end{tabular}
}
\end{table}

%Besides achieving the highest accuracy, our results demonstrate that AlexNet was \ac{CNN} that required the least time for training. Although the VGG-11 reached a high accuracy, this model demanded the most significant time in the training process, being unsuitable in \ac{FL} scenarios where the training time is essential.
%At the same time, the VGG-11 demanded the most significant amount of time in the training process and reached a low accuracy.

Our achieved accuracy is comparable to that found in the state of the art using the same parameters and dataset. Furthermore, our experiments suggest that ShuffleNet had the lowest accuracy when trained under the \ac{FL} paradigm, about $75.65\%$.

To analyze the relationship between the accuracy achieved and the training time, we applied the statistical relationship test between the variables Training Time and Accuracy. Thus, through the \textit{Pearson} correlation, we found that $r = -0.2$ leads us to infer that the relationship between training time and accuracy has a weak negative relationship. For our experiments, a weakly negative correlation means that the accuracy was high when training time was low, except for VGG-11. Also, the training time relates to the hyperparameters of each \ac{CNN}, such as learning rate, batch size, and others. Our measurements bring the rate of training time.

\begin{figure}[!htbp]
  \centering
  \includegraphics[width=1\textwidth]{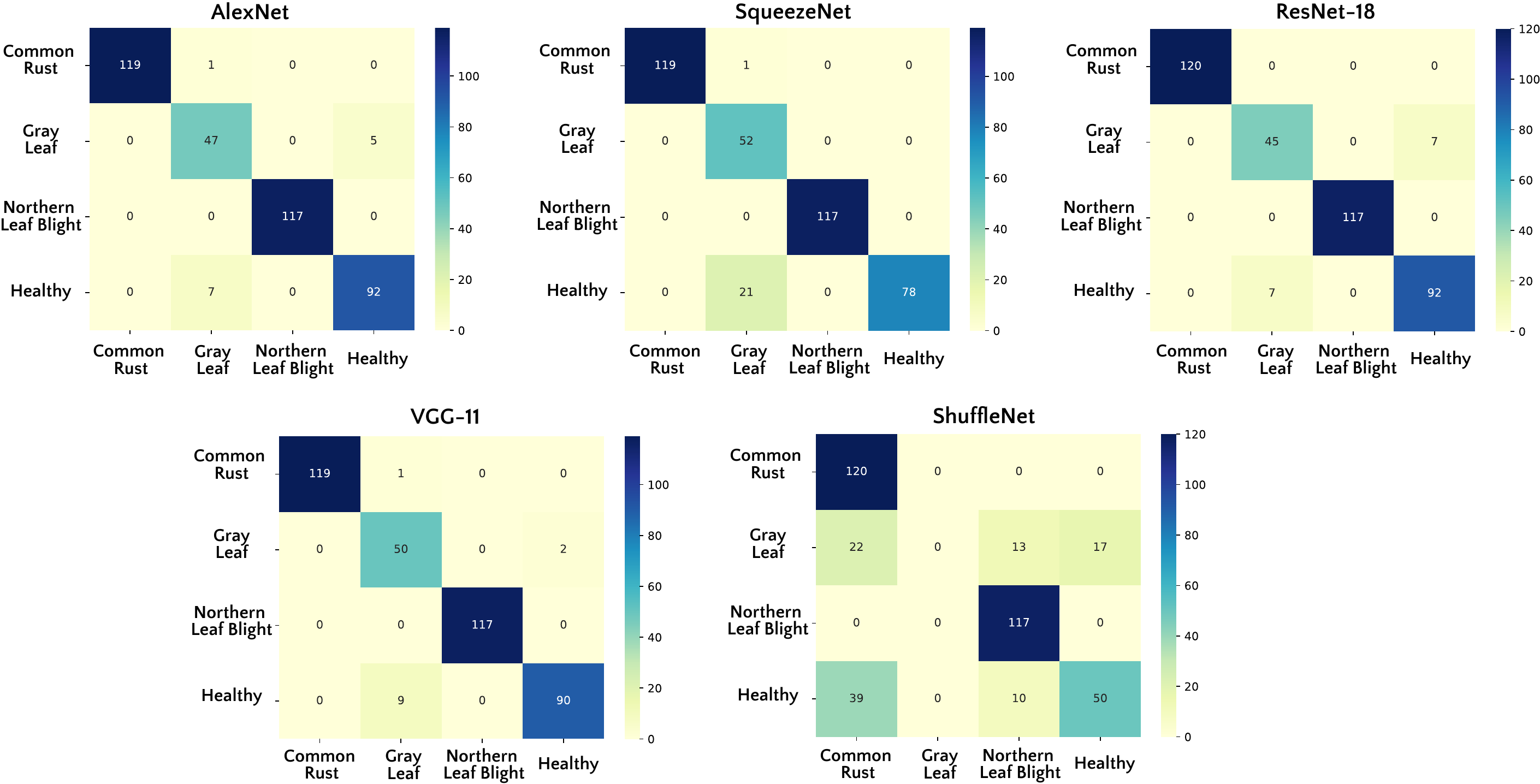}
  \caption{Confusion matrices for each tested CNN.}
  \label{fig:cm}
\end{figure}

In order to understand the accuracy of the models, we built confusion matrices to verify which classes of the dataset posed the most challenges in the training and generalization process of the model. We verified that for all \ac{CNNs} the northern leaf blight class was correctly classified. On the other hand, the gray leaf and healthy classes proved to be more challenging. In Figure~\ref{fig:cm} we consolidate the confusion matrices generated from the \ac{CNNs}.

Also, we evaluate the traffic volume used in the training process by the \ac{FL} technique. This traffic volume relies on all data transferred between the federated clients in the training process. We summarize these results in Table~\ref{tab:traffic_volume_by_cnn}, and it is worth noting that SqueezeNet, despite its classification performance, required less network traffic volume in the training process due to the number of \ac{CNN} trainable parameters. 

The results indicates that the traffic volume is related to the \ac{CNN} structure. CNNs with several layers and complex structures need to have the weights recorded in tensor data structures, generating large model sizes and, ultimately, high communication demands. In \ac{FL} framework, we must define the round of training, which refers to the number of times of server requires from federated clients the training process.

\begin{table}[!htbp]
\centering
\renewcommand{\arraystretch}{1}\scalefont{0.8}
\caption{Traffic Volume for each \ac{CNN}.}
\label{tab:traffic_volume_by_cnn}
\begin{tabular}{lcc}
\hline \hline
\multicolumn{1}{c}{\textbf{CNN}} & \textbf{Traffic Volume (MB)} & \textbf{Number of Parameters} \\ \hline
AlexNet                          & 84.24                   & 57,020,228                    \\
SqueezeNet                       & 17.70                   & 737,476                       \\
ResNet-18                        & 34.14                   & 11,178,564                    \\
VGG-11                           & 90.92                   & 128,788,228                   \\
ShuffleNet                       & 30.18                   & 1,257,704                     \\ \hline \hline
\end{tabular}
\end{table}

%SqueezeNet requires less traffic volume due to the number of \ac{CNN} trainable parameters. So, the traffic volume relates to the \ac{CNN} structure; some have many layers and structures where each weight of layers must be recorded in \ac{CNN} data structures, implying model size and ultimately to the communication demands. In \ac{FL} framework, we must define the round of training, which refers to the number of times of server requires from federated clients the training process.
The built federated learning rationale enhances data privacy trained where it is generated. In our carried experiments, we consider that training the maize leaf dataset in different and distributed clients makes it possible to assume that each one will only share the model weights and not the data each one has.

\section{Concluding Remarks}\label{sec:conclusion}

Machine learning techniques have proved suitable for several tasks, such as disease prediction based on image of agricultural crops. In this paper, we evaluate the potential of \ac{FL} to mitigate the data privacy gap that centralized machine learning techniques have. Thus, we evaluated the applicability of the \ac{FL} method through five \ac{CNNs} using maize leaf disease classification. For this, we built a short state-of-the-art survey. We found that the application of \ac{CNNs} to predict diseases in maize leaves predominantly did not consider the distributed paradigm and, consequently, did not support privacy. 

%Our numerical results suggest that AlexNet, due to its structure, is suitable to operate in the \ac{FL} model. Furthermore, it was possible to perceive a weak negative correlation between accuracy and training time, leading us to conclude that among the models considered, in the distributed training approach, the training time is inversely proportional to the accuracy of the models. Also, through numerical results, we measured and concluded that in the distributed learning paradigm, the number of \ac{CNN} parameters strongly correlates with the volume of data exchanged in the training process. 

Our numerical results suggest that AlexNet, due to its structure, lower training time, and high accuracy, is suitable to operate in the \ac{FL} model. Furthermore, it was worth noticing a weak negative correlation between accuracy and training time, leading us to conclude that for most of the models considered, in the distributed training approach, the training time is inversely proportional to the accuracy of the models. Also, through numerical results, we measured and concluded that the number of \ac{CNN} parameters in the distributed learning paradigm strongly correlates with the volume of data exchanged in the training process.

Also, our numerical and practical evaluation indicated that \ac{FL} could potentially enhance data privacy in heterogeneous domains. For future work, urges to evaluate alternatives weight aggregation techniques of \ac{CNNs} in addition to verifying the impact of network failures on the model convergence time.

{\small{\section*{Acknowledgment}
The authors acknowledge the financial support of the Federal University of Viçosa. Also, this study was financed in part by the Coordenação de Aperfeiçoamento de Pessoal de Nível Superior – Brasil (CAPES) – Finance Code 001.}}

{\small{\bibliographystyle{sbc}
\bibliography{sbc-template.bib}}}

%\bibliographystyle{sbc}
%\bibliography{sbc-template.bib}

\end{document}